%% file: main.tex
\documentclass{article}

\usepackage{microtype}
\usepackage{graphicx}
\usepackage{subfigure}
\usepackage{booktabs}
\usepackage{hyperref}
\usepackage{xcolor}
\usepackage{bbm}
\usepackage{multirow}
\usepackage{array}
\usepackage{threeparttable}
\usepackage{caption}

\usepackage{graphicx}

\usepackage[accepted]{icml2025}

\usepackage{amsmath}
\usepackage{amssymb}
\usepackage{mathtools}
\usepackage{amsthm}
\usepackage{enumitem}
\usepackage[capitalize,noabbrev]{cleveref}

\makeatletter
\renewcommand{\ICML@appearing}{}
\makeatother
\theoremstyle{plain}

\theoremstyle{definition}

\theoremstyle{remark}

\usepackage[textsize=tiny]{todonotes}

\usepackage{float}

\icmltitlerunning{Step-by-step Teacher for Sparsely Annotated Oriented Object Detection}

\begin{document}

\twocolumn[
\icmltitle{S$^2$Teacher: Step-by-step Teacher for \\ Sparsely Annotated Oriented Object Detection}
\icmlsetsymbol{corresponding}{\dag}

\begin{icmlauthorlist}
\icmlauthor{Yu Lin}{xmu}
\icmlauthor{Jianghang Lin}{xmu}
\icmlauthor{Kai Ye}{xmu}
\icmlauthor{You Shen}{xmu}
\icmlauthor{Yan Zhang}{xmu}
\icmlauthor{Shengchuan Zhang}{xmu}
\icmlauthor{Liujuan Cao}{xmu,corresponding}
\icmlauthor{Rongrong Ji}{xmu}
\end{icmlauthorlist}

\icmlaffiliation{xmu}{School of Informatics, Xiamen University, Xiamen, China}

\icmlcorrespondingauthor{Liujuan Cao}{caoliujuan@xmu.edu.cn}

\icmlkeywords{Machine Learning, ICML}

\vskip 0.3in
]

\printAffiliationsAndNotice{}  
\input{sec/0_abstract}
\input{sec/1_introduction}
\input{sec/2_related_work}
\input{sec/3_method}
\input{sec/4_experiments}
\input{sec/5_conclusion}


\clearpage
\twocolumn[]
\bibliography{example_paper}
\bibliographystyle{icml2025}

\newpage
\appendix
\section*{Appendix}
\section{More implementation details.}
Our implementation is based on rotated FCOS \cite{ref14}, using ResNet50 pretrained on ImageNet as the backbone. We use SGD optimizer with momentum set to 0.9, weight decay set to 0.0001, batch size set to 4, and learning rate adjustment strategy following \cite{ref5}. In terms of SAOOD, we use weak data augmentation for the teacher model and strong data augmentation for the student model. Weak augmentation includes random flipping, while strong augmentation includes random flipping, color jitter, random grayscale, and random Gaussian blur. Random flipping includes horizontal, vertical, and diagonal, with the random probability all set to 0.25. Following the previous teacher student model framework, we set the momentum of EMA to 0.9996 and update the teacher model using EMA after each iteration.

\section{More detailed comparative experiments with semi-supervised methods.}
We compare our method with state-of-the-art semi-supervised oriented object detection approaches on the DOTA-v1.5 dataset. Notably, the definition of k\% differs across settings. In semi-supervised learning, k\% means that k\% of the images are fully annotated, while the rest are unlabeled. In SAOOD, however, k\% means that k\% of the instances in each image are annotated. To fairly compare annotation costs, we define the Box Ratio (see Equation \ref{eq_appendix}). Since the total number of objects is same, a lower Box Ratio implies fewer annotated boxes and thus lower annotation cost. As shown in Table \ref{Tab_appendix1}, S$^2$Teacher achieves an mAP of 59.59\% at a Box Ratio of 5.5\%, clearly outperforming the semi-supervised method at 10\% Box Ratio. When the Box Ratio increases to 6.0\%, S$^2$Teacher reaches an mAP of 62.05\%, surpassing the semi-supervised methods at 19.6\%. At 8.3\% Box Ratio, S$^2$Teacher further improves to 63.13\%, exceeding the 62.63\% mAP achieved by the advanced semi-supervised method at a much higher Box Ratio of 32.6\%. In other words, S$^2$Teacher delivers better performance with only a quarter of the annotation cost. This is mainly because semi-supervised methods focus annotation cost on a subset of images is unworthy, as many objects in the same remote sensing images share similar features. As a result, high performance requires more annotated images, and thus a higher label cost. In contrast, SAOOD distributes annotation costs across more images, enabling the model to learn diverse features and avoid overfitting. S$^2$Teacher further improves this by using the learned sparse annotated features to step-by-step minning pseudo labels for similar objects. This allows it to achieve near fully-supervised performance with only 8.3\% of the annotation cost. These results suggest that S$^2$Teacher is better suited for achieving high detection performance at extremely low annotation cost in remote sensing images with densely annotated scenes.
\begin{equation}
    \text{Box Ratio} = \frac{N_A}{N_{\text{total}}},
    \label{eq_appendix}
\end{equation}
where $N_A$ means the number of annotated objects, and $N_{\text{total}}$ means the total number of objects in the training set.

\begin{table*}[t]
  \centering
  \caption{Detailed comparative experiments with semi-supervised methods on DOTA-v1.5.}
  \vspace{5pt}
  \resizebox{0.9\textwidth}{!}{
    \begin{tabular}{ccccc}
    \toprule
    Annotation method & Method & \multicolumn{3}{c}{mAP(\%)} \\
    \midrule
    \multirow{2}[4]{*}{fully-supervised} & \multirow{2}[4]{*}{Baseline (Rotated FCOS)} & \multicolumn{3}{c}{Box Ratio(100\%)} \\
\cmidrule{3-5}          &       & \multicolumn{3}{c}{64.42} \\
    \midrule
    \multirow{8}[4]{*}{Semi-supervised} &       & Box Ratio(10.0\%) & Box Ratio(19.6\%) & Box Ratio(32.6\%) \\
\cmidrule{3-5}          & SOOD \cite{ref5} (Rotated FCOS-based) & 48.63 & 55.58 & 59.23 \\
          & SOOD++ \cite{sood_p} (Rotated FCOS-based) & 50.48 & 57.44 & 61.51 \\
          & Dual Teacher \cite{dual_teacher} (Rotated FCOS-based) & 55.7  & 59.2  & 59.3 \\
          & Dual Teacher (ConvNext-based) & 55.6  & 60.7  & 62.0  \\
          & S$^2$O-Det \cite{S2O-Det} (Rotated FCOS-based) & 52.56 & 59.08 & 61.60  \\
          & S$^2$O-Det (ATSS-based) & 52.41 & 59.90  & 62.58 \\
          & MCL \cite{MCL} (Rotated FCOS-based) & 52.98 & 59.63 & 62.63 \\
    \midrule
    \multirow{2}[4]{*}{SAOOD} &       & Box Ratio(5.5\%) & Box Ratio(6.0\%) & Box Ratio(8.3\%) \\
\cmidrule{3-5}          & S$^2$Teacher(Rotated FCOS-based) & \textbf{59.59} & \textbf{62.05} & \textbf{63.13} \\
    \bottomrule
    \end{tabular}
    }
  \label{Tab_appendix1}
  \vspace{-10pt}
\end{table*}

\section{Performance limit exploration experiment.}
\label{section_appendix2}
We also explored the performance limits of S$^2$Teacher. Due to some small objects (such as cars) unlabeled in DOTA-v1.0, and to avoid unfair evaluation caused by S$^2$Teacher detecting these objects during testing, we explored the performance limits of S$^2$Teacher on the more challenging DOTA-v1.5. As shown in Table \ref{Tab_appendix2} in the appendix, the detection accuracy of S$^2$Teacher improves with higher annotation ratios. At 20\%, the mAP reaches saturation at 65.23\%, surpassing the fully-supervised rotated FCOS (64.42\%) and outperforming other state-of-the-art weakly or semi-supervised methods (e.g., SOOD \cite{ref5} only achieved 59.23\% mAP with 30\% labeled data, \cite{weakly_semi_sup} achieved mAP of 60.17\% with 30\% image full annotation, and 70\% image point annotation). This is mainly because some objects are manually missed in the DOTA-v1.5 dataset, which, when used for model training, can cause the detector to confuse foreground and background features, similar to misleading negative samples. S$^2$Teacher is able to mine these missed objects and add them as pseudo labels for training, thereby achieving better performance. Moreover, S$^2$Teacher continuously mines new and more challenging pseudo labels for training, which helps model with regularization and prevents the model from overfitting to the training set. This also demonstrates that S$^2$Teacher performs well on datasets with numerous small objects, achieving higher accuracy at lower annotation costs. Notably, the improvement in mAP is more significant with lower annotation ratios. Therefore, if we balance detection accuracy and annotation efficiency, 5\% annotation ratio may be a better choice.
\begin{table}[t]
  \centering
  \caption{Exploring performance limits experiment on DOTA-v1.5.}
  \vspace{5pt}
  \resizebox{0.9\columnwidth}{!}{
    \begin{tabular}{ccc}
    \toprule
    Annotation ratio & Method & mAP(\%) \\
    \midrule
    \multirow{2}[2]{*}{1\%} & Baseline (Rotated FCOS) & 51.50 \\
          & S$^2$Teacher & 59.59 \\
    \midrule
    \multirow{2}[2]{*}{2\%} & Baseline & 55.03 \\
          & S$^2$Teacher & 62.05 \\
    \midrule
    \multirow{2}[2]{*}{5\%} & Baseline & 56.75 \\
          & S$^2$Teacher & 63.13 \\
    \midrule
    \multirow{2}[2]{*}{10\%} & Baseline & 59.37 \\
          & S$^2$Teacher & 63.52 \\
    \midrule
    \multirow{2}[2]{*}{20\%} & Baseline & 62.14 \\
          & S$^2$Teacher & \textbf{65.23} \\
    \midrule
    \multirow{2}[2]{*}{30\%} & Baseline & 61.88 \\
          & S$^2$Teacher & 64.78 \\
    \bottomrule
    \end{tabular}%
    }
  \label{Tab_appendix2}%
  \vspace{-20pt}
\end{table}%

\section{More analysis on Gaussian modeling of information entropy.}
\begin{figure}[t]
    \begin{center}
        \centerline{\includegraphics[width=\columnwidth]{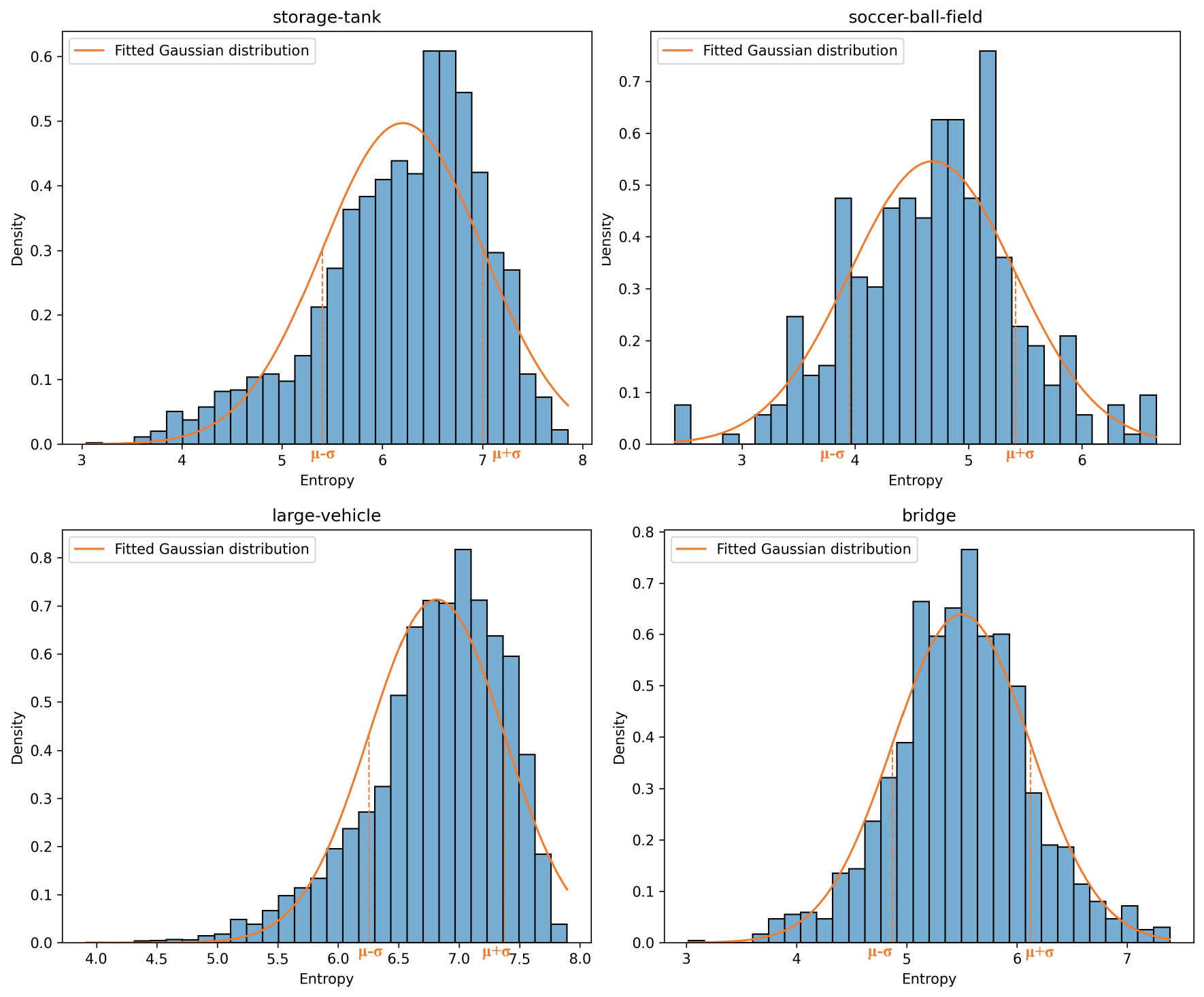}}
        \vspace{-5pt}
        \caption{Histogram and fitted Gaussian distribution of object information entropy on DOTA dataset.}
        \label{Fig_appendix_add1}
    \end{center}
    \vspace{-20pt}
\end{figure}

The complexity of image regions is often positively correlated with information entropy, which measures the uncertainty in pixel distribution, reflecting the diversity and complexity of textures \cite{active}. Foreground regions, due to their intricate structures, textures, and edge variations, typically exhibit a more diverse pixel distribution, resulting in higher information entropy. In contrast, background areas are usually smoother or more uniform, leading to lower entropy. Additionally, different object categories exhibit distinct information entropy distributions due to variations in texture structures. For example, large vehicles, with their complex details and textures, generally have higher information entropy, while soccer-ball-field, with more regular and monotonous textures, show lower entropy. Many natural phenomena, such as image noise and gradients, follow a Gaussian distribution. For objects within the same category, due to the large number of image samples and random texture variations, the entropy values across these samples can be treated as independent and identically distributed random variables. According to the Central Limit Theorem (CLT), when entropy measurements are taken from a sufficient number of image regions, their statistical distribution will approximate a Gaussian distribution. Therefore, we modeled the entropy of same-category objects using a Gaussian distribution. As shown in Figure \ref{Fig_appendix_add1}, we visualized the histograms of entropy distributions for different objects in the DOTA dataset. The results show that the entropy distributions of same-category objects closely follow a Gaussian distribution. Although some objects may exhibit slight peak shifts due to factors such as lighting conditions, resulting in a skewed distribution, most of the data still fall within the range of $[\mu \pm \sigma]$.

\begin{figure*}[t]
    \begin{center}
        \includegraphics[width=0.85\textwidth]{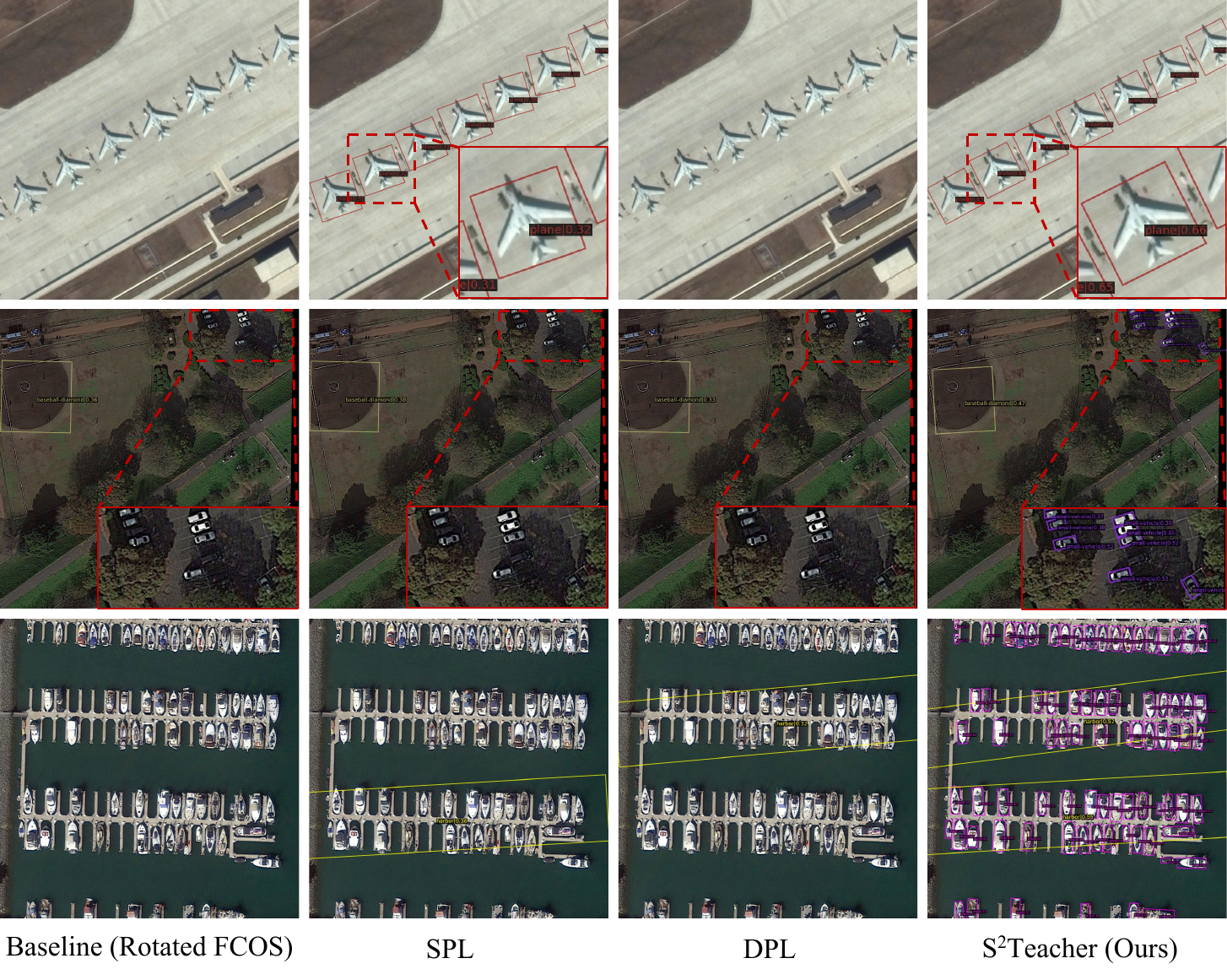}
        \vspace{-5pt}
        \caption{Visualization of the detection results of different methods trained on 10\% annotated DOTA-v1.0.}
        \label{Fig_appendix1}
    \end{center}
    \vspace{-10pt}
\end{figure*}
\section{Visual analysis of test results.}
We visualized the detection results of different pseudo-label generation methods based on teacher-student models. Existing pseudo-label generation methods can be broadly categorized into two types based on pseudo-label sparsity: sparse pseudo labels (SPL) and dense pseudo labels (DPL) \cite{ref5}. SPL \cite{ref18, dual_teacher} typically select teacher predictions after post-processing (e.g., score thresholds and NMS) to generate pseudo-labels, while DPL \cite{ref20, sood_p} bypass NMS and directly adopt dense outputs (e.g., post-sigmoid logits) from the teacher. As shown in Figure \ref{Fig_appendix1}, when trained on 10\% annotated DOTA-v1.0, the baseline fails to detect the object due to confusion between positive and negative sample features. This is mainly because sparse annotating causes unlabeled objects to be assigned as negative samples during training, making it difficult for the detector to distinguish between foreground and background features, leading to detection failure. In contrast, the other two teacher-student-based pseudo label generation methods alleviate this issue by mining unlabeled objects as pseudo labels and incorporating them into training. However, the confidence of their predicted remains low. For example, the SPL method yields a confidence of only about 0.3 for airplanes, indicating that the detector has not fully learned the features of airplanes. Benefiting from Focal Ignore Loss (reduces interference from misleading samples) and the high-quality pseudo label mining of S$^2$Teacher (enriches foreground representation while avoiding excessive pseudo label noise), S$^2$Teacher not only detects all objects but also shows significantly improved confidence, with a confidence level of 0.65 or higher for airplanes. Additionally, S$^2$Teacher performs equally well in detecting small objects, such as cars under tree shade. For dense scenes, like the ships docked at the port, S$^2$Teacher also performs effectively.
\end{document}

%% file: sec/0_abstract.tex
\begin{abstract}
Although fully-supervised oriented object detection has made significant progress in multimodal remote sensing image understanding, it comes at the cost of labor-intensive annotation. Recent studies have explored weakly and semi-supervised learning to alleviate this burden. However, these methods overlook the difficulties posed by dense annotations in complex remote sensing scenes. In this paper, we introduce a novel setting called sparsely annotated oriented object detection (SAOOD), which only labels partial instances, and propose a solution to address its challenges. Specifically, we focus on two key issues in the setting: \textbf{(1) sparse labeling leading to overfitting on limited foreground representations}, and \textbf{(2) unlabeled objects (false negatives) confusing feature learning}. To this end, we propose the S$^2$Teacher, a novel method that progressively mines pseudo-labels for unlabeled objects, from easy to hard, to enhance foreground representations. Additionally, it reweights the loss of unlabeled objects to mitigate their impact during training. Extensive experiments demonstrate that S$^2$Teacher not only significantly improves detector performance across different sparse annotation levels but also achieves near-fully-supervised performance on the DOTA dataset with only 10\% annotation instances, effectively balancing detection accuracy with annotation efficiency. The code will be public.
\end{abstract}

%% file: sec/1_introduction.tex
\section{Introduction}
\label{sec1}
Remote sensing images are usually collected from different sensors, thus exhibiting multimodal characteristics. The rapid interpretation of information in complex remote sensing images is of great practical value.
Oriented object detection has achieved great success in understanding the remote sensing images in recent years \cite{kfiou, gwd, oriented_detection}.
Remote sensing images often contain directional information due to overhead view.
Unlike horizontal detectors, oriented detectors predict rotated boxes to capture this information and enable accurate localization.
%
%
A key challenge hindering the development of oriented object detection is the high annotation cost, as labeling a rotated box (RBox) is approximately 36.5\% more expensive than labeling a horizontal box (HBox) \cite{ref1}. 

\input{figures/comparison}
To address this, recent studies have explored training oriented detectors using HBox supervision~\cite{ref1, ref2}, point supervision~\cite{ref3, ref4}, and semi-supervision~\cite{ref5} to reduce annotation costs.
%
These methods have yielded promising results in oriented object detection.
%
However, they overlook a key characteristic of remote sensing images: \textbf{the prevalence of densely labeled scenes}.
%
Studies have shown that dense scenes can significantly activate the amygdala of the human brain~\cite{ref6}, making it more prone to fatigue. 
As shown in Figure \ref{Fig1}, the small size, partial occlusion, and blurred features of objects in remote sensing images make it \textbf{extremely challenging to label all objects without omission}.
This requires annotators to repeatedly check to prevent missing objects, which is time-consuming.
%
In fact, due to the challenges associated with annotating dense, small objects (such as tightly packed parked cars), many cars in the widely used DOTA-v1.0~\cite{ref7} dataset are unlabeled.
%
Even after relabeling in DOTA-v1.5, missing labels remain (see Section \ref{sec4_5}).

%
As the saying goes, ``Birds of a feather flock together.'' 
Under a macroscopic remote sensing view, objects that exhibit spatial clustering are typically of the same class with similar features.
Partial annotation of such objects can capture most of their features.
%
As shown in Figure\,\ref{Fig1}, we conducted experiments on the DOTA dataset. For a dense scene image containing 411 instances, annotating rotated boxes (RBox) requires about 2708s, annotating horizontal boxes (HBox) takes about 1713s, and point annotation takes around 1381s. In contrast, sparse annotation of only 10\% of the instances requires just 306s.
%
%
Although HBox and point annotations reduce labeling time by 36.7\% and 49\% respectively, their efficiency gains are limited due to the time-consuming process of checking missed objects in dense scenes. In contrast, sparse annotation achieves a higher efficiency gain (88.7\%) by not only reducing the number of labeled instances but also eliminating the need for exhaustive checking.
%
Based on this observation, we naturally introduced a novel setting to balance detection accuracy and annotation efficiency: sparsely annotated oriented object detection (SAOOD), which annotated partial instances.
\textbf{The key advantage of SAOOD is}: 1) Compared to weakly and semi-supervised methods, SAOOD enables randomly instance annotation in dense scenes, eliminating the need for repeatedly check to avoid missing objects, which is time-consuming. 2) It also avoids the misleading supervision and harm training caused by label-omitted objects due to indistinct features in remote sensing images.  3) Semi-supervised methods concentrate annotation costs on a subset of image, which is unworthy, as most instance features in the same image are similar. SAOOD distributes annotation costs across more images, allowing the model to see a richer feature distribution and effectively prevent overfitting.
Pseudo-label generation is a commonly used method in semi supervised object detection~\cite{unbiased, ref20}, and an intuitive idea is to apply it to SAOOD.
However, directly applying pseudo-label generation to SAOOD will arise two main issues: 1) Objects that are not annotated in the image are treated as negative samples during training. Since their features resemble those of positive samples, they will mislead gradients and confuse detectors. 2) The limited number of labeled objects leads to insufficient positive features, increasing the risk of overfitting. Most SSOD methods generate pseudo-labels for all instances in the image at once. However, when feature learning is insufficient, these pseudo-labels often introduce significant noise, misleading detector and limiting overall performance.
%
%
To address these issues, we propose a progressive pseudo-label generation framework, called S$^2$Teacher.
%
It clusters the Top-k highest confidence proposals to obtain pseudo labels, filters out false positive (FP) pseudo labels using information entropy Gaussian modeling, gradually freezes high-confidence pseudo labels unchanged through multi-temporal comparisons, and prompts S$^2$Teacher to mining more hard pseudo labels. This easy-to-hard pseudo-label mining method steadily improves detector performance, avoiding excessive pseudo-label noise. During training, our Focal Ignore Loss mitigates the impact of unlabeled objects by reweighting the loss, preventing misleading negative samples from confusing the detector.
%
%
%
Just as teachers progressively introduce concepts from simple to complex, we refer to this method as the step-by-step teacher (S$^2$Teacher).
Our main contributions are as follows:
\vspace{-5pt}
\begin{itemize}[left=0pt]
    \item We analyze a key factor for the high annotation cost in remote sensing: the dense annotation scenario, and discuss how SAOOD can address this. We propose a new teacher-student framework called S$^2$Teacher for SAOOD, which improves the performance of oriented detectors through step-by-step pseudo-label mining.
    \item We propose a novel pseudo-label generation method, utilizing Top-k high-confidence proposal clustering and information entropy Gaussian modeling to mine unlabeled objects. This addresses the issue of limited foreground representation while minimizing pseudo-label noise. By gradually freezing pseudo labels, the teacher model is encouraged to mine unlabeled objects from easy to hard. Our Focal Ignore Loss mitigates the impact of unlabeled objects misleading the training by reweighting loss.
    \item Extensive experiments show that compared to other state-of-the-art methods, S$^2$Teacher not only achieves higher detection performance with lower annotation costs, but also achieves near fully-supervised performance on the DOTA dataset with only 10\% annotation.
\end{itemize}

%% file: figures/comparison.tex

\begin{figure}[t]
    \vskip 0.2in
    \begin{center}
        \centerline{\includegraphics[width=0.95\columnwidth]{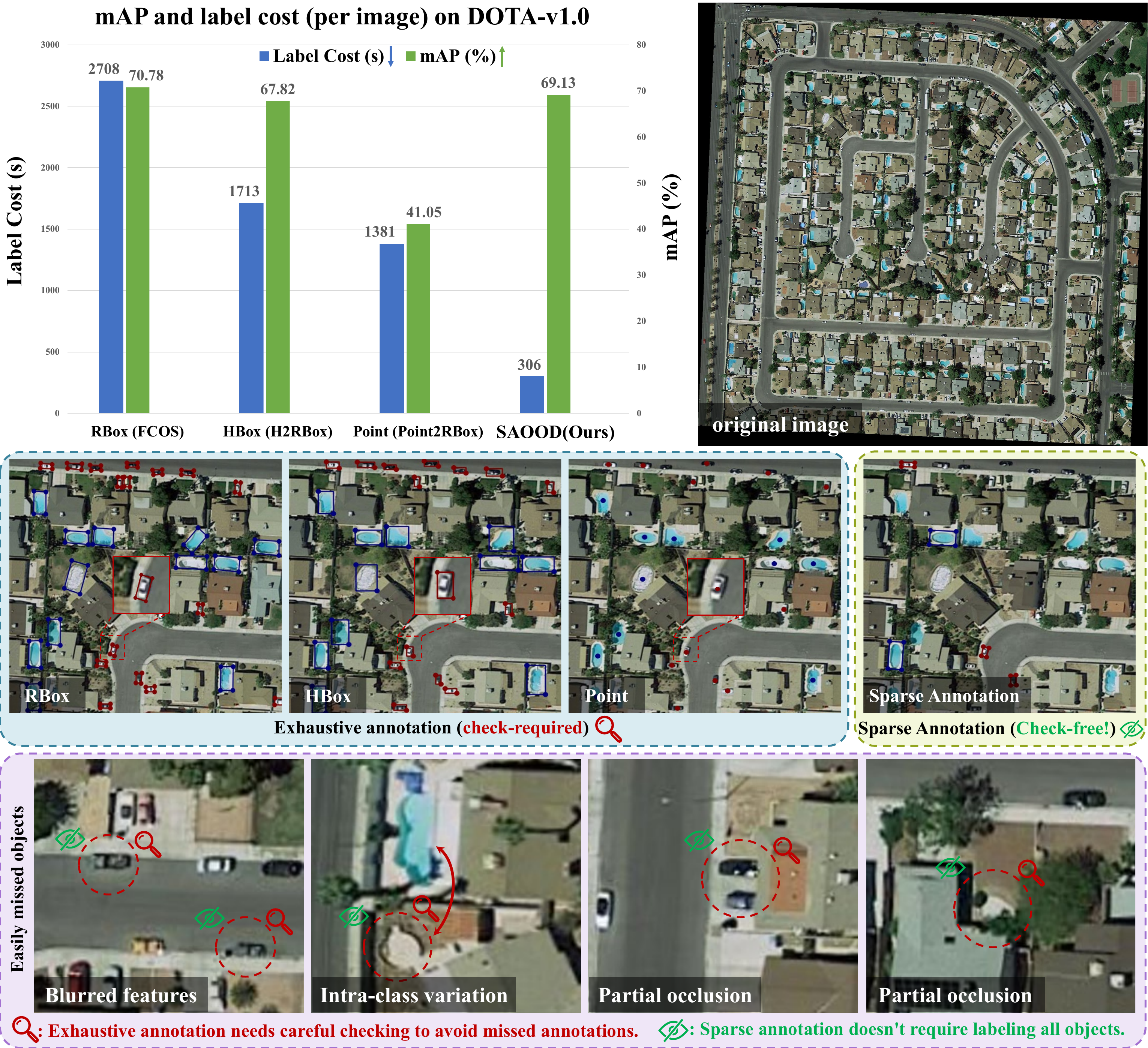}}
        \vspace{-5pt}
        \caption{Compare different annotation methods. RBox (full supervision), HBox, and point supervision require labeling all objects and careful checking to avoid missed annotations, which is time-consuming. In remote sensing, dense small objects and issues like blurring and occlusion make labeling all objects difficult. Sparse annotation randomly labels partial objects without check, greatly reducing cost. Our S$^2$Teacher approaches full supervision performance under this setting.}
        \label{Fig1}
    \end{center}
    \vskip -30pt
\end{figure}

%% file: sec/2_related_work.tex
\section{Related Work}
\textbf{Fully-supervised oriented object detection.} Oriented object detection has been widely applied in remote sensing \cite{ref8, ref9}, scene text \cite{ref10}, and retail \cite{ref11} in recent years. Typical oriented detectors include two-stage detector Oriented R-CNN \cite{ref12}, one-stage detector S$^2$A-Net \cite{ref13}, and anchor-free detector Rotated FCOS \cite{ref14}. Previous studies have focused on improving the performance of oriented detectors through model improving, such as feature alignment \cite{ref15}, addressing angular boundary issue \cite{ref16}, and using large convolution kernel \cite{ref17}. With the rise of the large models, researchers have recognized the importance of data-driven approaches for model performance. 
%
However, in remote sensing, the small and dense objects make labeling all instances time-consuming and challenging.
%
\input{figures/framework}
\textbf{Weakly and semi-supervised oriented object detection.}
%
Recently, studies have begun to focus on reducing the annotation cost of oriented object detection to obtain more training data.
They can be mainly divided into three types: 1) weakly supervised approaches: these methods reduce annotation costs by converting RBox annotations into weaker forms such as HBox annotations \cite{ref1} or point annotations \cite{ref3, ref4, Pointobb-v2}. While HBox-supervised oriented detectors achieve higher accuracy, the reduction in annotation cost is limited. Point-supervised oriented detectors significantly reduce annotation costs, but their detection accuracy is far inferior to fully-supervised detectors. 2) Semi-supervised approaches: Some methods \cite{ref5, dual_teacher, MCL} introduce semi-supervised methods into oriented object detection. These methods can balance detection accuracy and annotation cost, but it does not consider the dense annotation problem of remote sensing images. 3) Weakly and semi-supervised combination approaches: Some methods \cite{weakly_semi_sup} use a combination of RBox and point annotations to reduce annotation costs while ensuring detection accuracy. However, they also do not consider the dense annotation issue, and the combination of annotations also limited the cost reduction of annotation. This paper solves the problem of dense annotation by introducing SAOOD. Sparse annotated object detection (SAOD) and semi-supervised object detection (SSOD) are related tasks, differing in that images in SSOD are either fully labeled or unlabeled, while images in SAOD are only partially instances labeled \cite{co_mining}. In natural scene images, where object numbers are small and annotating all instances in an image is easier, SAOD receives less attention compared to SSOD. However, in remote sensing, where scenes are dense (e.g., DOTA-v1.5 has an average of 143 instances per image compared to 7 in COCO \cite{ref7}), detailed annotation all instances is labor-intensive, making SAOOD a more elegant way to reduce annotation costs.

\textbf{Sparsely annotated object detection.} Recent studies have explored sparsely annotated object detection (SAOD) in natural scenes. Co-mining \cite{co_mining} employs a collaborative mining mechanism with a Siamese network, where two branches generate pseudo labels for each other to discover unlabeled instances. Region-based \cite{ref25} reformulate the task as a region-level semi-supervised problem, identifying regions likely to contain unlabeled objects. Calibrated Teacher \cite{ref24} enhances confidence calibration, aligning score distributions across detectors and overcoming the limitations of fixed thresholds, achieving state-of-the-art results. However, these approaches degrade in remote sensing scenarios, where objects are denser and unlabeled instances are far more abundant than in natural scenes. This reduces the quality of pseudo labels generated by these methods. To address this, our S$^2$Teacher adopts step-by-step pseudo-label mining to ensure label reliability in dense scenes, and design Focal Ignore Loss to reduce the effects of numerous unlabeled objects during training.

%% file: figures/framework.tex
\begin{figure*}[t]
    \begin{center}
        \centerline{\includegraphics[width=\textwidth]{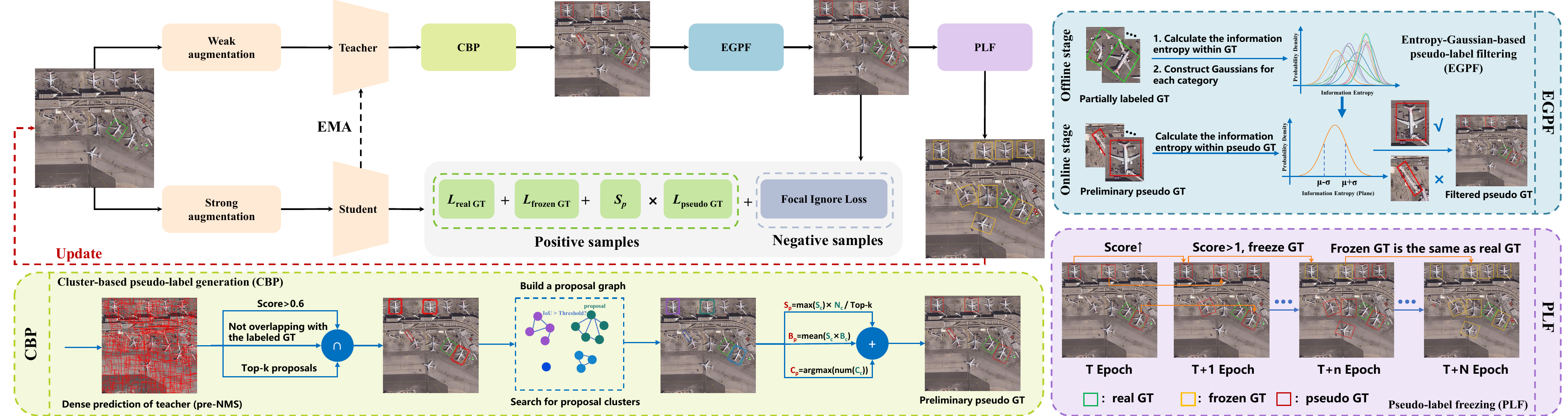}}
        \caption{The overall framework of the S$^2$Teacher. The input image is processed through teacher model and CBP to prioritize the mining of easy unlabeled objects. After filtering out false positives through the EGPF, the pseudo GT is used for training the student model. The pseudo GT mined by each iteration are compared through the PLF, gradually freezing high confidence pseudo GT, prompting the CBP to continuously mine harder unlabeled objects.}
        \label{Fig2}
    \end{center}
    \vskip -20pt
\end{figure*}

%% file: sec/3_method.tex
\section{S$^2$Teacher}
Given training images $\mathcal{X}$, $\mathcal{Y}_s$ represents the sparse annotated instance set, and $\mathcal{Y}_u$ is the unlabeled instance set. 
The goal of SAOOD is to use $\left\{ \mathcal{X}, \mathcal{Y}_s \right\}$ to initially train a model, mine pseudo labels $\mathcal{Y}_p$ from $\mathcal{Y}_u$, and then continue training with $\left\{ \mathcal{X}, \mathcal{Y}_s \cup \mathcal{Y}_p \right\}$, thereby iteratively improving performance. 
The overall structure of S$^2$Teacher is shown in Figure \ref{Fig2}, which is built upon the classic teacher-student model framework. 
The input image is weakly enhanced for the teacher model and strongly enhanced for the student model. 
The teacher model utilizes our proposed cluster-based pseudo-label generation module (CBP) to generate preliminary pseudo-labels in the image. 
These pseudo-labels then pass through the pseudo-label filtering module based on information entropy Gaussian modeling (EGPF), resulting in high-quality pseudo-labels that are used for supervised training of the student model.
%
In each iteration, the pseudo-labels are sequentially compared using the pseudo-label freezing module (PLF), gradually freezing high-confidence pseudo-labels as real labels. 
This forces the teacher model to continue mining new, difficult pseudo-labels.
The training loss of student models is divided into two parts: positive sample loss and negative sample loss. 
Positive sample loss includes real ground truth (GT), frozen GT, and pseudo GT loss. 
The negative sample loss adopts our designed Focal Ignore Loss.
\subsection{Cluster-based Pseudo-label Generation}
A common method for generating pseudo labels in Semi-Supervised Object Detection (SSOD) is to filter the predictions of the teacher model using a classification confidence threshold, followed by Non-Maximum Suppression (NMS) to obtain the final labels \cite{ref18}. 
However, when the teacher's predictions are erroneous, the resulting pseudo-labels may mislead the student model.
Although techniques such as confidence-weighted losses \cite{ref19} and the use of logits as soft labels for supervised training \cite{ref20} have reduced the risk of misleading false positive (FP) pseudo-labels, they have generally overlooked the role of group decision-making. 
%
Specifically, individual predictions are subject to randomness, but when multiple proposals near a particular location exhibit high classification confidence, it is highly likely that these correspond to potential unlabeled objects.
%
Additionally, previous pseudo-label generation methods generate all possible pseudo-labels for the entire image in each iteration. 
We argue that this approach is too aggressive, as the limited number of labeled objects constrains the feature learning of the teacher model. 
%
Generating all pseudo-labels at once introduces substantial noise, misleading the student model. 
In oriented object detection, this issue is exacerbated by the added angle dimension, which expands the solution space and increases the likelihood of erroneous predictions
%
Therefore, we propose generating only the Top-k proposals with the highest confidence to create pseudo-labels. 
%
After training the model with these pseudo-labels to enhance feature learning and performance, more difficult pseudo-labels can be mined. 
This step-by-step training gradually improves detection accuracy.
%

To address the aforementioned issues, we proposed a cluster-based pseudo-label generation module (CBP). 
As shown in Figure\,\ref{Fig2}, the weakly enhanced image is passed through the teacher model to obtain the pre-NMS output, which is then fed into the CBP. 
%
The CBP filters proposals based on three criteria: 1) applying a confidence threshold to eliminate low-score background proposals; 2) removing proposals with a high IoU with real ground truth (GT) (since objects with real GT in SAOOD images do not require pseudo-labels); and 3) selecting the Top-k proposals with the highest confidence from the remaining proposals.
%
The filtered proposals are further used to construct a proposal graph. Inspired by \cite{ref21}, we treat each proposal as a node and calculate the IoU between each proposal.
Two proposals with an IoU greater than a threshold (0.5) are considered connected, thus forming a proposal graph.
The interconnected proposal nodes in this graph form clusters, and we use a greedy algorithm to search for each cluster.
The position of each proposal cluster is treated as indicative of a potential unlabeled object.
%
We calculate the cluster score $S_p$ by considering both the number of proposals within the cluster and the classification score (more proposals suggest a higher likelihood of potential unlabeled objects), which is formulated as:
%
\begin{equation}
    {S_p} = \mathop {\max }\limits_{i \in [1,{N_c}]} ({S_{c,i}}) \cdot \frac{{{N_c}}}{k},
    \label{eq1}
\end{equation}
where $S_{c,i}$ is the classification score of the $i$-th proposal in the proposal cluster, and $N_c$ is the number of proposals in each cluster.
We use $S_p$ as the confidence of pseudo GT. 
%
Then, we calculate the weighted average position of the proposals within the cluster to obtain the pseudo GT bounding box $B_p$:
\begin{equation}
    {B_p}(x,y,w,h,\theta ) = \frac{1}{{{N_c}}}\sum\limits_{i = 1}^{{N_c}} {{S_{c,i}} \cdot {B_{c,i}}(} x,y,w,h,\theta ),
    \label{eq2}
\end{equation}
where $B_{c,i}$ is that of the $i$-th proposal in the proposal cluster.
%
Finally, the category of the pseudo GT is taken to be the category with the most proposals within the cluster:
\begin{equation}
    {C_p} = \mathop {\arg \max }\limits_{c \in \{ 1,2, \cdots ,N\} } \sum\limits_{i = 1}^{{N_c}} {\mathbbm{1}_{\{c_i = c\}}},
    \label{eq3}
\end{equation}
where $\mathbbm{1}$ is the indicator function, being 1 if $c_i = c$ and 0 otherwise, $c$ is the category, and $c_i$ is the category of the $i$-th proposal in the cluster.
This group decision-making approach avoids the randomness of individual proposals and improves the confidence of the pseudo-labels.
\subsection{Pseudo-label Filtering}
A significant issue with pseudo-label-based methods is the interference of false positive (FP) pseudo labels during training.
Due to inevitable prediction errors in the teacher model, FP pseudo-labels can mislead the gradient direction of the student model.
This issue is especially prominent in SAOOD because the objects are only sparsely annotated.
%
To mitigate this, we use Focal Ignore Loss (refer to Section \ref{sec3_4} for details) to reduce the weight of potential targets (negative samples) in the loss function.
However, reducing the weight of negative samples also increases the occurrence of FP pseudo-labels, which limits the performance of pseudo-label-based methods in SAOOD.
To address this issue, we designed a pseudo-label filtering module based on information entropy Gaussian modeling (EGPF), as illustrated in Figure \ref{Fig2}. 
Information entropy is commonly used to measure the complexity and diversity of image content \cite{active}.
%
As shown in Formula \ref{eq4}, assuming the probability distribution of each pixel in an image region be denoted as $\{p_1, p_2, \dots, p_n\}$. When all probabilities are equal (i.e., \( p_1 = p_2 = \dots = p_n = \frac{1}{n} \)), the entropy reaches its maximum value \( \log n \). Conversely, when a single probability \( p_i = 1 \), the entropy becomes zero. Background regions are typically smooth, with simple content and minimal pixel variation, resulting in pixel values concentrated around a few values and hence lower entropy. In contrast, foreground regions tend to exhibit more complex structures, greater pixel variation, and a more dispersed distribution of pixels, leading to higher entropy.
We define object information entropy $\cal H$ as:
\begin{equation}
    {\cal H} =  - \sum\limits_{i = 1}^n p ({x_i})\log (p({x_i})),
    \label{eq4}
\end{equation}
where $n$ is the number of pixels in the box, and $p({x_i})$ is the value distribution of each pixel in the box.
%
Due to differences in material, texture, color, and other visual characteristics, the entropy of objects varies across categories. Objects with complex visual content, such as ships, tend to exhibit higher entropy, while those with relatively simple appearance, such as sports fields, generally show lower entropy.
Consequently, the EGPF first calculates $\cal H$ of all manually annotated objects within the real GT and constructs a Gaussian distribution $p({{\cal H}_c})$ of $\cal H$ for each object category:
\begin{equation}
    p({{\cal H}_c}) = \frac{1}{{\sqrt {2\pi \sigma _c^2} }}\exp \left( { - \frac{{{{({{\cal H}_c} - {\mu _c})}^2}}}{{2\sigma _c^2}}} \right),
    \label{eq5}
\end{equation}
where $\mu _c$ is the mean of ${{\cal H}_c}$, and $\sigma _c^2$ is the variance of ${{\cal H}_c}$.
%
This step is performed automatically before training on a new dataset and only adds time during the first training phase, without affecting inference time.
At each iteration, EGPF calculates ${{\cal H}_{pseudo}}$ of objects within the pseudo GT mined by the CBP, and removes the FP pseudo labels by: 
\begin{equation}
    {\rm{Filter}}({{\cal H}_{pseudo}}) = \mathbbm{1}{_{\{ \mu  - \sigma  \le {{\cal H}_{pseudo}} \le \mu  + \sigma \} }},
    \label{eq6}
\end{equation}
where $\rm{Filter}$ represents filtering pseudo GT, $\mathbbm{1}$ is the indicator function. When $\mu  - \sigma  \le {{\cal H}_{pseudo}} \le \mu  + \sigma$ is present, the current pseudo GT is retained, otherwise it is filtered out.
%
This filtering process ensures the quality of pseudo-labels and prevents a large number of FP pseudo-labels from misleading the student model.
\subsection{Pseudo-label Freezing}
To ensure high-quality pseudo-labels, CBP selects the Top-k proposals with the highest scores for pseudo-label mining. 
%
However, this may lead to overlapping pseudo-labels in each iteration, preventing the teacher model from discovering new unlabeled objects.
%
Therefore, we propose the Pseudo-label Freezing Module (PLF), which complements the CBP to enables teacher model mine more difficult pseudo-labels.
The structure of PLF is shown in Figure\,\ref{Fig2}. The pseudo-labels mined in each iteration are stored in a queue, with the queue length corresponding to the number of iterations per epoch.
PLF calculates the IoU between the pseudo GT mined in the current epoch and those from the previous epoch to determine if they correspond to the same object.
%
When a pseudo GT is mined multiple times at the same location, it indicates a high probability of an unlabeled object in that region.
In this case, PLF increases the confidence of the pseudo GT.
%
Conversely, when a pseudo GT is mined at a location in previous epochs but not in the current epoch, it suggests a decreased likelihood of an unlabeled object, prompting PLF to reduce the confidence of that pseudo GT.
After each iteration, the mined pseudo-labels are stored in the queue following the first-in, first-out (FIFO) principle.
If a location is repeatedly mined for pseudo GT, its confidence continues to increase.
Once the confidence exceeds 1, it indicates a high probability of an unlabeled object at that location.
%
At this point, PLF freezes the pseudo GT as a real GT, which is treated as true ground truth in the loss calculation and does not require further mining. 
%
This process allows the CBP to focus on mining pseudo-labels for new unlabeled objects.
%
In essence, PLF serves as a temporal group decision mechanism, using pseudo-label mining results from multiple epochs to jointly determine the likelihood of an unlabeled object at a given location.
\subsection{The overall loss}
\label{sec3_4}
\textbf{Total Loss.} The overall loss function of S$^2$Teacher consists of positive and negative sample losses.
The positive loss includes the loss from manually annotated real GT and the pseudo GT generated by the teacher model.
The pseudo GT loss is divided into two components: the frozen pseudo GT, which is treated with the same weight (1.0) as real GT, and the ordinary pseudo GT, which is weighted by $S_p$ from the CBP. 
%
The negative sample loss is using Focal Ignore Loss $\mathcal{L}_{\text{F-I}}^{cls}$.
Therefore, the total loss can be formulated as:
\begin{align}
    {\cal L}_{\text{total}} &= \sum\limits_{i \in \text{pos}} \left( 
    {\cal L}_{\text{GT},i} \cdot \mathbbm{1}_{\{ i \in \text{GT} \}} + 
    {\cal L}_{\text{frz GT},i} \cdot \mathbbm{1}_{\{ i \in \text{frz GT} \}} \right. \notag \\
    &\quad \left. + S_p \cdot {\cal L}_{\text{pseu GT},i} \cdot \mathbbm{1}_{\{ i \in \text{pseu GT} \}} \right) + \sum\limits_{j \in \text{neg}} \mathcal{L}_{\text{F-I}}^{cls},
    \label{eq7}
\end{align}
where ${\cal L}_{\text{GT},i}$ is the loss of real GT, ${\cal L}_{\text{frz GT},i}$ is the loss of frozen GT, ${\cal L}_{\text{pseu GT},i}$ is the loss of pseudo GT, all of them are composed of classification and regression losses, Focal Loss \cite{ref22} is used for classification, and IoU loss \cite{iou} is used for regression. $\mathbbm{1}$ is the indicator function.
When the $i$-th proposal is assigned to the real GT, it is 1, otherwise it is 0, and the same applies to others. 
\textbf{Focal Ignore Loss.} As shown in Figure \ref{Fig3}, in SAOOD, objects are partially labeled, and proposals around unlabeled objects are treated as negative samples  (misleading samples) during training, despite sharing the same features with positive samples.
This mislabeling misguides the detector, causing it to confuse foreground and background features. 
The one-stage, anchor-free detector typically considers all proposals that do not intersect with GT as negative samples, leading to far more misleading samples than positive ones and misleading the gradient direction.
Additionally, these detectors rely on Focal Loss, which treats misleading samples as hard negatives, further exacerbating the issue.
\input{figures/pseudo_labels}
To address this issue, we designed Focal Ignore Loss, which can focus on truly hard negative samples while ignoring misleading samples.
The formula for Focal Ignore Loss is shown in \ref{eq8}, for misleading negative samples, their features resemble those of positive samples, causing the teacher model to predict a relatively low background confidence for them.
Therefore, using it as the loss weight can avoid numerous misleading samples dominating training. 
It is worth noting that the background confidence of hard negative samples around real GT is also relatively low. 
If the weights of hard negative samples are also reduced, it will prevent the model from learning their features, resulting in over prediction of the foreground. 
Therefore, we calculate the IoU between proposals with low background confidence and real GT, distinguishing the loss calculation of hard negative samples with high IoU from misleading negative samples. 
This approach prevents the down-weighting of hard negative samples from causing numerous false positives (FP).
\begin{align}
    {{\cal L}_{\text{F-I}}^{cls}} =& - \frac{1}{{{N_{hn}}}}\sum\limits_{n = 1}^{{N_{hn}}} {\sum\limits_{i = 1}^C {{\alpha _i}{{(1 - p_{n,i}^S)}^\gamma }\log (p_{n,i}^S)} } \notag - \frac{1}{N_n} \\  & \cdot \sum_{m=1}^{N_n} (1 - q_m^T) \sum_{i=1}^{C} \alpha_i \left( 1 - p_{m,i}^S \right)^\gamma \log \left( p_{m,i}^S \right),
    \label{eq8}
\end{align}
where $N_{hn}$ is the number of hard negative samples, $C$ is the number of categories, $\alpha_i$ and $\gamma$ are hyperparameters set the same as Focal Loss \cite{ref22}, $p_{n,i}^S$ is the probability predicted by student model that the n-th sample belongs to the actual class, $N_n$ is the number of normal negative samples, and $q_m^T$ is the probability predicted by teacher model that the m-th sample belongs to the foreground.

%% file: figures/pseudo_labels.tex
\begin{figure}[t]
    \vskip 0.1in
    \centering
    \begin{minipage}{0.45\columnwidth}
        \centering
        \includegraphics[width=\columnwidth]{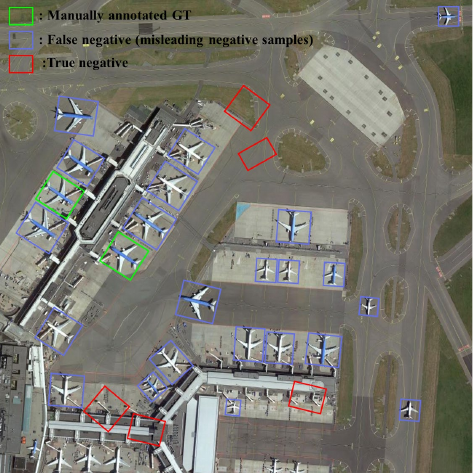}  
        \vspace{-20pt}
        \caption[Numerous false negatives (FN) mislead training.]{Numerous false negatives mislead training.}
        \label{Fig3}
    \end{minipage}
    \hspace{0.005\textwidth}  
    \begin{minipage}{0.45\columnwidth}
        \centering
        \includegraphics[width=\columnwidth]{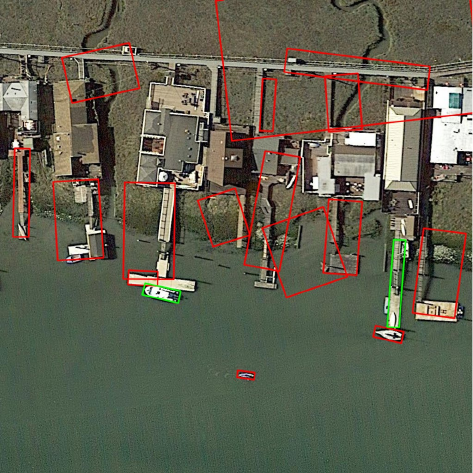}  
        \vspace{-20pt}
        \caption[Prior methods often generate FP pseudo-labels.]{Prior methods generate FP pseudo-labels.}
        \label{Fig4}
    \end{minipage}
    \vskip -12pt
\end{figure}

%% file: sec/4_experiments.tex
\section{Experiments}
\subsection{Datasets}
We conducted experiments on the widely used remote sensing datasets DOTA-v1.0 \cite{ref7} and DOTA-v1.5. As this is a SAOOD task, sparse annotated datasets must be created by sampling labels. Previous SAOD method \cite{ref23} generated sparse datasets by selecting a fixed proportion of samples from each category in the complete dataset. This method assumes the annotator knows the number of samples for each category in the complete dataset, but in practice, the annotator lacks this prior information, leading to discrepancies between the actual data distribution and ideal sampling. We suggest using an alternative sampling method \cite{ref24, ref25}. It randomly samples objects by category within each images. When the product of sampling ratio and the number of objects results in a non-integer, the value is rounded, and at least one object is retained to prevent zero-shot issue. This method better reflects the real-world scenario, as the annotator can see the number of different objects in the image and label more instances of the classes with higher counts. Due to the large size of DOTA dataset images, we followed the previous work \cite{ref5, ref15, ref26} and cropped them into 1024 $\times$ 1024 patches. For model evaluation, we used mean average precision (mAP) with an IoU threshold of 0.5.

\textbf{DOTA-v1.0.} DOTA-v1.0 includes 2806 aerial images, 15 categories. These categories are defined as: Plane (PL), Baseball Diamond (BD), Bridge (BR), Ground Track Field (GTF), Small Vehicle (SV), Large Vehicle (LV), Ship (SH), Tennis Court (TC), Basketball Court (BC), Storage Tank (ST), Soccer-Ball Field (SBF), Roundabout (RA), Harbor (HA), Swimming Pool (SP), and Helicopter (HC).

\textbf{DOTA-v1.5.} DOTA-v1.5 adds many extremely small instances (such as cars), and adds a new category: Container Crane (CC).

\subsection{Experimental settings}
Following the pseudo-label generation paradigm of SOOD \cite{ref5}, \textbf{we use Rotated FCOS as the baseline, which only uses sparse annotated instances to train the model}. The implementation and hyperparameter settings are the same as those in mmrotate \cite{ref26}. We use weak data augmentation for the teacher model and strong augmentation for the student model. All models were trained on 4 RTX3090 GPUs using SGD optimizer, with an initial learning rate set to 0.0025, momentum set to 0.9, weight decay set to 0.0001. Following the previous teacher-student model, we update the teacher model parameters using EMA with momentum set to 0.9996.

\subsection{Main result}

\textbf{Comparisons with the state-of-the-arts.} We first compared the detection accuracy of state-of-the-art methods with different annotation approaches on DOTA-v1.0. Notably, the definition of the split ratio varies across settings: semi-supervised methods annotate k\% of images, while SAOOD annotates k\% of instances. To ensure a fair comparison of annotation costs, we introduce the Box Ratio, defined in Equation \ref{eq9}. For example, in Table \ref{Tab_add1}, 7.9\% RBox indicates that 7.9\% of objects in the train set are annotated with RBox, while 100\% Point indicates that all objects in the train set are annotated with points. Since the total number of objects in the training set is equal, a higher Box Ratio means more objects are annotated, thereby implying a higher annotation cost when the annotation form is the same.
\begin{equation}
    \text{Box Ratio} = \frac{N_A}{N_{\text{total}}},
    \label{eq9}
\end{equation}
where $N_A$ means the number of annotated objects, and $N_{\text{total}}$ means the total number of objects in the training set.
As shown in Table \ref{Tab_add1}, H2RBox achieves 67.82\% mAP, whereas Figure \ref{Fig1} shows that the annotation cost of HBox remains relatively high. Point supervision further reduces labeling cost but suffers a significant drop in accuracy, with PointOBB-v2 reaching only 44.85\% mAP. The performance of semi-supervised methods strongly depends on the number of annotated images: S$^2$O-Det achieves 55.18\% mAP with 10\% annotated images, and 67.70\% with 30\%. This requires a trade-off between accuracy and annotation cost. In contrast, under a similar labeling cost (Box Ratio of 7.9\%), our S$^2$Teacher achieves an mAP of 64.59\%, significantly outperforming S$^2$O-Det (55.18\%). When the Box Ratio increases to 14\%, S$^2$Teacher achieves 69.13\% mAP, even surpassing the 67.70\% of S$^2$O-Det with a 27.6\% Box Ratio. In other words, S$^2$Teacher attains superior mAP at nearly half the annotation cost. We believe that the semi-supervised method of concentrating annotation costs on a subset of images is unworthy, as most object features in the same remote sensing image are similar. SAOOD distributes the annotation costs across more images, enabling the model to learn a broader feature distribution. Building on this, S$^2$Teacher further mines similar objects as pseudo labels, which are added to training and boost performance. Additionally, as shown in Figure \ref{Fig1}, SAOOD is suitable for dense annotation scenes in remote sensing images, avoiding repeated checks for missed objects and greatly reducing annotation costs.

\input{tables/different_annotation}
\input{tables/sota_dota1.0}
\textbf{Results on DOTA-v1.0.} As shown in Table \ref{Tab1}, our S$^2$Teacher significantly outperforms the baseline across various annotation ratios. With 1\% annotation ratio, the baseline one-stage detector, Rotated FCOS, achieves 57.17\% mAP, while S$^2$Teacher improves it to 64.59\%. This gain is attributed to its ability to leverage learned features from sparse annotations to mine unlabeled objects as pseudo labels from easy to hard, forming a self-improving cycle. At 10\% annotation ratio, S$^2$Teacher (Rotated FCOS-based) reaches 69.13\% mAP, approaching the fully-supervised Rotated FCOS (70.78\%). Notably, S$^2$Teacher shows significant gains in detecting densely packed objects, such as large vehicles (LV) and small vehicles (SV), as well as objects with distinct features like basketball courts (BC) and storage tanks (ST), which highlights its effectiveness in reducing the high annotation cost in dense scenarios. For example, in remote sensing images where large vehicles are often densely distributed, the baseline mAP for LV under 1\% supervision is only 50.9\%, while S$^2$Teacher increases it to 66.6\%, marking a gain of 15.7\%. Moreover, S$^2$Teacher is also compatible with two-stage detectors, showing significant performance gains over the baseline when applied to Oriented R-CNN across various annotation ratios.
\input{tables/sota_dota1.5}
\input{tables/saood_compared_experiments}
\textbf{Compared with other teacher-student methods on DOTA-v1.5.} We also compare S$^2$Teacher, the progressive pseudo-labeling generating approach, with other teacher-student-based pseudo label generation methods. Existing methods can be broadly categorized into two types based on pseudo-label sparsity: sparse pseudo labels (SPL) and dense pseudo labels (DPL) \cite{ref5}. SPL \cite{ref18, dual_teacher} typically select teacher predictions after post-processing (e.g., score thresholds and NMS) to generate pseudo-labels, while DPL \cite{ref20, sood_p} bypass NMS and directly adopt dense outputs (e.g., post-sigmoid logits) from the teacher. As shown in Table~\ref{Tab_add_dota15}, both SPL and DPL offer limited gains under the SAOOD setting, as they attempt to mine all unlabeled objects at once. Under sparse annotation, since numerous unlabeled objects confusing the foreground feature learning, mining all unlabeled objects at once can easily generate numerous FP pseudo labels (as shown in Figure \ref{Fig4}). In contrast, S$^2$Teacher adopts a progressive strategy, gradually mining unlabeled objects from easy to hard. The model first mines and learns from high-confidence, easily recognizable unlabeled objects (e.g., sports fields in Figure~\ref{Fig6}), as its capacity improves, it will discover and learn harder instances (e.g., cars in Figure~\ref{Fig6}). This step-wise process enhances learning stability, reduces noise, and performs better under sparse supervision.

\input{figures/visualization}
\textbf{Compared with other sparse annotation methods.} We compare our method with existing sparse annotated object detection (SAOD) approaches on DOTA-v1.0 under 5\% annotation. As shown in Table \ref{Tab_add_saod}, our S$^2$Teacher achieves the highest mAP, with particularly significant improvement for one-stage detectors. This is because remote sensing scenes contain denser and smaller objects, making natural scene SAOD methods less effective due to weak object features and blurred class boundaries, which lead to many FP pseudo-labels and limiting performance. S$^2$Teacher alleviates this by gradually mining unlabeled objects from easy to hard and ensure pseudo-label quality through proposal cluster-based group decision-making. Additionally, two-stage detectors inherently perform better under sparse annotations, as they sample negatives in the RPN stage, effectively filtering out many unlabeled objects. In contrast, one-stage detectors treat all proposals not overlapping with ground truth as negatives, causing many unlabeled objects being misused as negative samples and confusing learning—a problem further exacerbated in remote sensing due to numerous unlabeled objects. Our Focal Ignore Loss can reduce the interference of unlabeled objects on training through loss reweighting, filling the gap between one-stage and two-stage detector in SAOOD.

\subsection{Ablation Studies}
We conducted ablation studies on several modules of S$^2$Teacher using 10\% annotated DOTA-v1.0. As shown in Table \ref{Tab3}, applying Focal Ignore Loss to the baseline yields only a limited improvement of 0.48\%, primarily because it can only reduces the influence of misleading samples but does not address the limited foreground representation. When combined with CBP, the mAP improves significantly by 4.76\%, as CBP continuously mines pseudo labels to train the student model and enhance foreground representation. Meanwhile, Focal Ignore Loss further suppresses misleading samples impact, promoting the mining of CBP. Incorporating EGPF raises mAP to 68.87\%, mainly due to its ability to filter out FP pseudo labels, thereby improving label quality and preventing them from misleading student model. Finally, adding PLF further boosts mAP to 69.13\%. PLF discovers pseudo labels across different iterations and gradually freezes those with high confidence. This temporal group decision-making ensures the quality of frozen labels and encourages CBP to explore new, harder samples, enabling step-by-step learning and continuous performance improvement.
\input{tables/ablation_modules}
\input{tables/ablatin_hyp_CBP}
\subsection{Hyperparameter experiment}
We conducted experiments on the CBP hyperparameters. As shown in Table \ref{Tab_add2}, when the score threshold and Top-k are set differently, there is little change in mAP (both within 1\%), indicating that CBP is not sensitive to hyperparameter settings. When the score threshold is set to 0.6 and Top-k is set to 30, mAP reaches its maximum.

\subsection{Visualization Analysis}
\label{sec4_5}
To more intuitively understanding the S$^2$Teacher, we visualized the pseudo labels mined. As shown in Figure \ref{Fig6_a}, our S$^2$Teacher can accurately mine unlabeled objects (orange boxes) and use them as pseudo-labels to train the student model, addressing the issue of insufficient foreground representation. Figure \ref{Fig6_c} shows the mining results of the same image at different iterations. We can see that the PLF gradually froze the pseudo-labels that are mined multiple times, forced the CBP to mining new unlabeled objects.
%
As shown in Figure \ref{Fig6_b}, we observed an interesting phenomenon during visualization. Due to the numerous objects and unclear features of some annotated objects (e.g., bridges and roundabouts in Figure \ref{Fig6_b}), the DOTA dataset contains many manually omitted instances, which reflects the annotation challenges in aerial datasets. However, S$^2$Teacher can discover these missed objects when mining pseudo-labels, which indirectly reflects the robustness of our method.

%% file: tables/different_annotation.tex
{
\begin{table}[t]
  \centering
  \caption{Comparison of state-of-the-art methods for different annotation methods on DOTA-v1.0. k\%$^\star$ means that k\% of the images are fully labeled. k\%$^\Delta$ means that k\% of the images are labeled with RBox, while the remaining are labeled with Point. $\lozenge$ means based on FCOS, $\ddagger$ (YOLOF), ${\clubsuit}$ (ReDet)}
  \vspace{5pt}
  \resizebox{\columnwidth}{!}{
    \begin{tabular}{cccc}
    \toprule
    Annotation method & Method & Box Ratio & mAP(\%) \\
    \midrule
    RBox-supervised & Rotated FCOS$^\lozenge$ \cite{ref14} & 100\% RBox & 70.78 \\
    \midrule
    HBox-supervised & H2RBox$^\lozenge$ \cite{ref1} & 100\% HBox & 67.82 \\
    \midrule
    \multirow{4}[0]{*}{Point-supervised} & Point2RBox$^\ddagger$ \cite{ref4} & \multirow{4}[0]{*}{100\% Point} & 41.05 \\
          & PointOBB$^\lozenge$ \cite{ref3} &       & 30.08 \\
          & PointOBB-v2$^\lozenge$ \cite{Pointobb-v2} &       & 41.68 \\
          & PointOBB-v2$^{\clubsuit}$ &       & 44.85 \\
    \midrule
    \multirow{4}[2]{*}{Semi-supervised} & \multicolumn{1}{c}{SOOD++$^\lozenge$ \cite{sood_p} (10\%$^\star$)} & \multirow{2}[1]{*}{7.9\% RBox} & 54.17 \\
          & S$^2$O-Det$^\lozenge$ \cite{S2O-Det} (10\%$^\star$) &       & 55.18 \\
\cmidrule{2-4}          & SOOD++$^\lozenge$ (30\%$^\star$) & \multirow{2}[1]{*}{27.6\% RBox} & 64.93 \\
          & S$^2$O-Det$^\lozenge$ (30\%$^\star$) &       & 67.70  \\
    \midrule
    \multirow{2}[0]{*}{Weakly semi-supervised} & Wu et al.\cite{weakly_semi_sup}$^\lozenge$ (10\%$^\Delta$) & 7.9\% RBox+92.1\% Point & 59.69 \\
          & Wu et al.\cite{weakly_semi_sup}$^\lozenge$ (30\%$^\Delta$) & 27.6\% RBox+72.4\% Point & 67.04 \\
    \midrule
    \multirow{2}[0]{*}{SAOOD} & \multirow{2}[0]{*}{S$^2$Teacher$^\lozenge$ (Ours)} & 7.9\% RBox & 64.59 \\
          &       & 14.0\% RBox & \textbf{69.13} \\
    \bottomrule
    \end{tabular}%
    }
  \label{Tab_add1}%
  \vspace{-10pt}
\end{table}%
}

%% file: tables/sota_dota1.0.tex
{
\begin{table*}[t]
  \centering
  \caption{The results of S$^2$Teacher based on one-stage and two-stage detectors on DOTA-v1.0 with various annotation ratios. $\star$ represents based on one-stage detector (Rotated FCOS \cite{ref14}), and $\Delta$ represents based on two-stage detector (Oriented R-CNN \cite{ref12}). Our S$^2$Teacher has significant performance improvements for different annotation ratios and detectors.}
  \vspace{5pt}
  \resizebox{0.95\textwidth}{!}{
    \begin{tabular}{cccccccccccccccccc}
    \toprule
    Annotation ratio & Method & PL    & BD    & BR    & GTF   & SV    & LV    & SH    & TC    & BC    & ST    & SBF   & RA    & HA    & SP    & HC    & mAP(\%) \\
    \midrule
    \multirow{4}[4]{*}{1\%} & Baseline (Rotated FCOS)$^\star$ & 76.0  & 72.9  & 36.5  & 60.8  & 43.0  & 50.9  & 68.2  & 86.9  & 57.7  & 49.4  & 55.9  & 63.6  & 45.7  & 40.8  & 49.2  & 57.17 \\
          & S$^2$Teacher (Rotated FCOS-based)$^\star$ & 86.0  & 79.0  & 39.1  & 58.9  & 54.9  & 66.6  & 70.7  & 90.3  & 74.2  & 58.4  & 61.1  & 63.6  & 58.0  & 51.1  & 57.2  & \textbf{64.59} \\
\cmidrule{2-18}          & Baseline (Oriented R-CNN)$^\Delta$ & 76.5  & 75.6  & 39.1  & 68.2  & 38.6  & 46.5  & 54.3  & 87.7  & 61.9  & 49.1  & 58.1  & 66.2  & 43.1  & 36.7  & 35.9  & 55.82 \\
          & S$^2$Teacher (Oriented R-CNN-based)$^\Delta$ & 84.3  & 79.1  & 40.0  & 72.5  & 58.1  & 72.6  & 84.4  & 89.9  & 72.9  & 60.3  & 61.8  & 63.4  & 61.9  & 45.3  & 46.7  & \textbf{66.21} \\
    \midrule
    \multirow{4}[4]{*}{2\%} & Rotated FCOS$^\star$ & 75.6  & 72.4  & 37.9  & 64.8  & 47.4  & 57.7  & 70.3  & 86.2  & 59.5  & 50.9  & 54.0  & 67.8  & 50.8  & 39.7  & 50.7  & 59.05 \\
          & S$^2$Teacher (Rotated FCOS-based)$^\star$ & 85.8  & 76.9  & 39.7  & 65.1  & 61.9  & 72.3  & 84.0  & 90.3  & 69.8  & 58.8  & 57.1  & 64.4  & 61.6  & 51.0  & 55.8  & \textbf{66.31} \\
\cmidrule{2-18}          & Oriented R-CNN$^\Delta$ & 75.8  & 74.7  & 41.9  & 71.8  & 45.2  & 55.0  & 63.1  & 87.8  & 60.9  & 48.3  & 58.6  & 66.1  & 43.1  & 31.7  & 37.6  & 57.43 \\
          & S$^2$Teacher (Oriented R-CNN-based)$^\Delta$ & 84.0  & 78.5  & 43.2  & 75.0  & 59.9  & 78.3  & 86.7  & 90.2  & 71.2  & 60.4  & 62.3  & 63.5  & 62.3  & 44.2  & 42.7  & \textbf{66.82} \\
    \midrule
    \multirow{4}[4]{*}{5\%} & Rotated FCOS$^\star$ & 81.4  & 71.6  & 33.6  & 62.4  & 56.8  & 60.0  & 78.5  & 86.6  & 58.6  & 52.5  & 56.7  & 63.5  & 43.6  & 44.0  & 51.8  & 60.11 \\
          & S$^2$Teacher (Rotated FCOS-based)$^\star$ & 86.3  & 77.9  & 39.7  & 62.5  & 65.7  & 77.8  & 87.0  & 90.4  & 73.6  & 60.2  & 58.8  & 62.2  & 60.7  & 53.7  & 57.7  & \textbf{67.62} \\
\cmidrule{2-18}          & Oriented R-CNN$^\Delta$ & 76.1  & 75.1  & 40.3  & 71.8  & 52.6  & 65.7  & 73.9  & 86.9  & 62.5  & 57.5  & 56.8  & 64.7  & 47.1  & 34.1  & 51.0  & 61.07 \\
          & S$^2$Teacher (Oriented R-CNN-based)$^\Delta$ & 86.1  & 74.7  & 46.0  & 75.2  & 66.5  & 81.3  & 88.1  & 90.2  & 76.4  & 61.4  & 63.8  & 66.7  & 63.5  & 48.3  & 46.7  & \textbf{68.98} \\
    \midrule
    \multirow{4}[4]{*}{10\%} & Rotated FCOS$^\star$ & 84.1  & 71.4  & 35.9  & 63.9  & 63.5  & 73.8  & 84.6  & 86.9  & 60.8  & 56.6  & 52.5  & 63.7  & 51.7  & 48.7  & 51.5  & 63.3 \\
          & S$^2$Teacher (Rotated FCOS-based)$^\star$ & 87.0  & 79.6  & 41.1  & 61.3  & 67.0  & 81.8  & 87.3  & 90.3  & 74.9  & 65.8  & 55.9  & 65.4  & 63.1  & 56.1  & 60.4  & \textbf{69.13} \\
\cmidrule{2-18}          & Oriented R-CNN$^\Delta$ & 77.7  & 73.2  & 42.9  & 71.6  & 58.3  & 71.6  & 84.0  & 87.3  & 61.8  & 59.4  & 58.1  & 69.6  & 53.5  & 44.1  & 51.2  & 64.29 \\
          & S$^2$Teacher (Oriented R-CNN-based)$^\Delta$ & 85.6  & 75.7  & 41.7  & 72.4  & 67.2  & 82.3  & 87.9  & 90.1  & 75.2  & 67.2  & 63.0  & 62.5  & 70.4  & 51.0  & 59.0  & \textbf{70.06} \\
    \bottomrule
    \end{tabular}%
    }
  \label{Tab1}%
  \vspace{-10pt}
\end{table*}%
}

%% file: tables/sota_dota1.5.tex

\begin{table}[t]
  \centering
  \caption{Compare different pseudo-label generation methods based on teacher-student frameworks on DOTA-v1.5.}
  \vspace{5pt}
  \resizebox{0.95\columnwidth}{!}{
    \begin{tabular}{c >{\centering\arraybackslash}p{4cm} >{\centering\arraybackslash}p{3cm}}
    \toprule
    Annotation ratio & Method & mAP(\%) \\
    \midrule
    \multirow{4}[2]{*}{1\%} & Baseline (Rotated FCOS) & 51.50 \\
          & SPL & 53.22 \\
          & DPL  & 51.78 \\
          & S$^2$Teacher(Ours) & \textbf{59.59} \\
    \midrule
    \multirow{4}[2]{*}{5\%} & Baseline (Rotated FCOS) & 56.75 \\
          & SPL & 56.83 \\
          & DPL & 57.01 \\
          & S$^2$Teacher(Ours) & \textbf{63.13} \\
    \midrule
    \multirow{4}[2]{*}{10\%} & Baseline (Rotated FCOS) & 59.37 \\
          & SPL & 59.59 \\
          & DPL & 59.59 \\
          & S$^2$Teacher(Ours) & \textbf{63.52} \\
    \bottomrule
    \end{tabular}%
    }
  \label{Tab_add_dota15}%
  \vspace{-10pt}
\end{table}%

%% file: tables/saood_compared_experiments.tex
{
\begin{table}[t]
  \centering
  \caption{Compare with other sparse annotation methods on the DOTA-v1.0 with 5\% sparse annotation. $\blacktriangle$ means based on one-stage detector, $\circ$ means based on two-stage detector.}
  \vspace{5pt}
  \resizebox{0.95\columnwidth}{!}{
    \begin{tabular}{ccc}
    \toprule
    Annotation method & Method & mAP(\%) \\
    \midrule
    \multirow{8}[1]{*}{Sparsely-annotated} & Calibrated Teacher$^\blacktriangle$ \cite{ref24} & 55.81 \\
          & Co-mining$^\circ$ \cite{co_mining} & 65.35 \\
          & Region-based$^\circ$ \cite{ref25} & 65.71 \\
          & PECL \cite{ref23} (S$^2$A-Net-based)$^\blacktriangle$ & 57.42 \\
          & PECL (ReDet-based)$^\circ$ & 67.06 \\
          & S$^2$Teacher (Rotated FCOS-based)$^\blacktriangle$ & 67.62 \\
          & S$^2$Teacher (Oriented R-CNN-based)$^\circ$ & \textbf{68.98} \\
    \bottomrule
    \end{tabular}%
    }
  \label{Tab_add_saod}%
  \vspace{-10pt}
\end{table}%
}

%% file: figures/visualization.tex
\begin{figure*}[t]
    \begin{center}
        \subfigure[Pseudo labels mined during the training process.]{
            \includegraphics[width=0.45\textwidth]{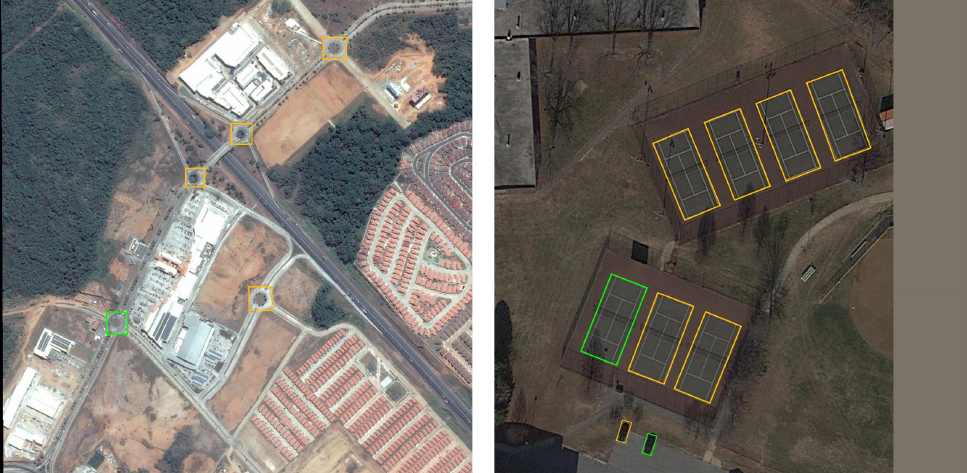}
            \label{Fig6_a}
        }
        \subfigure[S$^2$Teacher found manually missed objects in labeling.]{
            \includegraphics[width=0.45\textwidth]{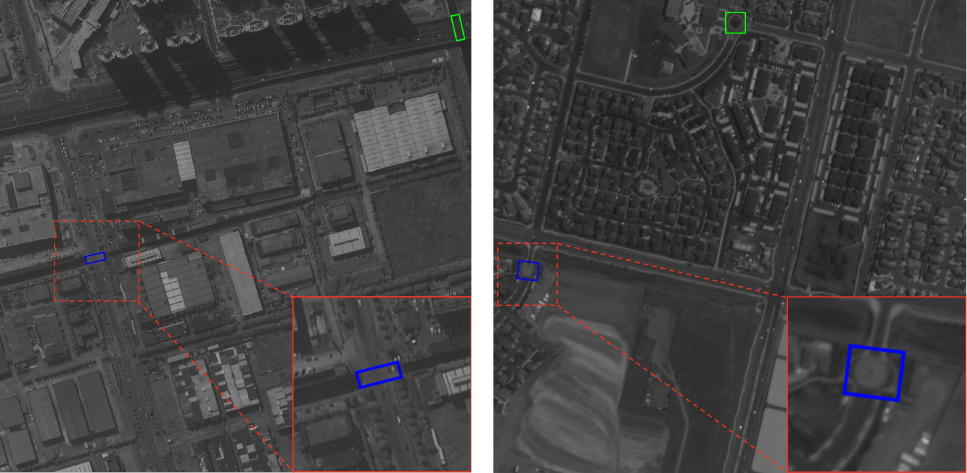}
            \label{Fig6_b}
        }
        \vskip -3pt
        \subfigure[Visualization of PLF freeze pseudo-label process in different iterations.]{
            \includegraphics[width=0.91\textwidth]{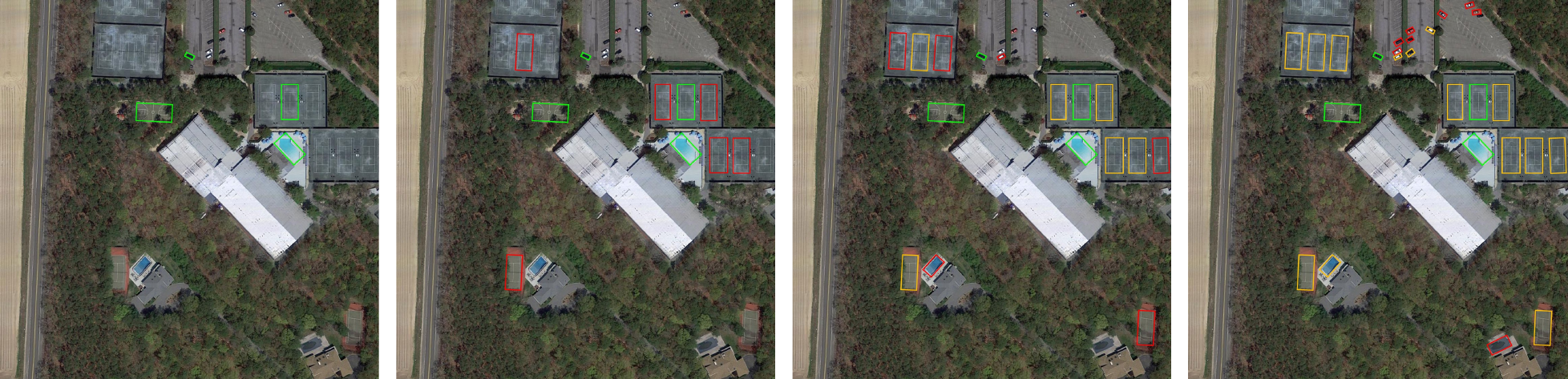}
            \label{Fig6_c}
        }
    \end{center}
    \vskip -15pt
    \caption{S$^2$Teacher pseudo label mining visualization. Among them, the \textcolor{green}{green} box is the manually annotated real GT, the \textcolor{red}{red} box is the pseudo GT, the \textcolor{orange}{orange} box is the frozen pseudo GT by PLF, and the \textcolor{blue}{blue} box is the mined pseudo GT, but it was missed during manual annotation, so it is mistakenly judged as FP.}
    \label{Fig6}
    \vspace{-10pt}
\end{figure*}

%% file: tables/ablation_modules.tex
\begin{table}[t]
  \centering
  \caption{The ablation study of each module in S$^2$Teacher.}
  \vspace{5pt}
  \resizebox{0.7\columnwidth}{!}{
    \begin{tabular}{ccccc}
    \toprule
    Focal Ignore Loss & CBP   & EGPF  & PLF   & mAP(\%) \\
    \midrule
          &       &       &       & 63.30  \\
    \checkmark     &       &       &       & 63.78 \\
    \checkmark     & \checkmark     &       &       & 68.06 \\
    \checkmark     & \checkmark     & \checkmark     &       & 68.87 \\
    \checkmark     & \checkmark     & \checkmark     & \checkmark     & \textbf{69.13} \\
    \bottomrule
    \end{tabular}%
    }
  \label{Tab3}%
  \vspace{-10pt}
\end{table}%

%% file: tables/ablatin_hyp_CBP.tex
\begin{table}[t]
  \centering
  \caption{Hyperparameter experiment in CBP.}
  \vspace{5pt}
  \resizebox{0.67\columnwidth}{!}{
    \begin{tabular}{cc|cc}
    \toprule
    Score threshold & mAP(\%) & Top-k & mAP(\%) \\
    \midrule
    0.5   & 68.26 & 10    & 68.96 \\
    \textbf{0.6}   & \textbf{69.13} & 20    & 68.65 \\
    0.7   & 68.79 & \textbf{30}    & \textbf{69.13} \\
    0.8   & 69.09 & 40    & 68.77 \\
    0.9   & 68.84 & 50    & 69.07 \\
    \bottomrule
    \end{tabular}%
    }
  \label{Tab_add2}%
  \vspace{-10pt}
\end{table}%

%% file: sec/5_conclusion.tex
\section{Conclusion}
We explore Sparsely Annotated Oriented Object Detection (SAOOD), a crucial yet underexplored task for reducing annotation costs in remote sensing images. To address the challenges of SAOOD, we propose S$^2$Teacher. By incrementally mining high-confidence pseudo labels, S$^2$Teacher mitigates the issue of limited foreground representation caused by sparse annotations. Additionally, Focal Ignore Loss minimizes the impact of misleading negative samples. Experimental results show that S$^2$Teacher achieves near fully-supervised performance with only 10\% annotated data, balancing detection accuracy and annotation efficiency.